\documentclass[10pt,twocolumn,letterpaper]{article}

\usepackage{wacv}
\usepackage{times}
\usepackage{epsfig}
\usepackage{graphicx}
\usepackage{amsmath}
\usepackage{amssymb}
\usepackage{multirow}
\usepackage{booktabs}
\usepackage{arydshln}

\usepackage[pagebackref=true,breaklinks=true,letterpaper=true,colorlinks,bookmarks=false]{hyperref}
\wacvfinalcopy
\def\wacvPaperID{***} 

\ifwacvfinal\fi
\begin{document}

\title{Towards a General Model of Knowledge for Facial Analysis\\ by Multi-Source  Transfer Learning}

\author{
	\begin{tabular*}{0.666\linewidth}{@{\extracolsep{\fill}}ccc}
		\multicolumn{3}{c}{Valentin Vielzeuf$^{1,2}$ ~~~ Alexis Lechervy$^{2}$ ~~~ St\'ephane Pateux$^1$ ~~~ Frederic~Jurie$^2$}
		\\
		\multicolumn{3}{c}{}\\
		\multicolumn{3}{c}{$^1$Orange Labs, Cesson-S\'evign\'e, France}\\
		\multicolumn{3}{c}{$^2$Universit\'e Caen Normandie, France}
	\end{tabular*}
}

\maketitle
\ifwacvfinal\thispagestyle{empty}\fi

\begin{abstract} 
This paper proposes a step toward obtaining   general models of knowledge for facial analysis, by addressing the question of multi-source transfer learning. More precisely, the proposed approach consists in two successive training steps: the first one consists in applying a combination operator to define a common embedding for the multiple sources materialized by different existing trained models. The proposed operator relies on an auto-encoder, trained on a large dataset, efficient both in terms of compression ratio and  transfer learning performance. In a second step we exploit a distillation approach to obtain a lightweight student model mimicking the collection of the fused existing models. This model outperforms its teacher on novel tasks, achieving results on par with state-of-the-art methods on 15 facial analysis tasks (and domains), at an affordable training cost. Moreover, this student has 75 times less parameters than the original teacher and can be applied to a variety of novel face-related tasks.
\end{abstract}

\section{Introduction}
An increasing number of deep neural networks has been implemented and trained during the last few years. These existing models can be seen as incredibly rich and compressed sources of knowledge about diverse domains, which can be reused to tackle novel tasks by transferring this knowledge. 
In this context, the standard way for knowledge transfer consists in selecting a single source, generally under the form of the parameters of a statistical model  (\eg, a pre-trained network), and to re-use it on a new task by fine-tuning the parameters. The knowledge source is often empirically chosen, typically by selecting the closest task according to human judgment or a complex rich task such as the ImageNet classification task. 

To automate this selection process, recent works~\cite{li2019task,achille2019task2vec,zamir2018taskonomy} have shown that a relational space between diverse basic tasks/models can be exploited, allowing to combine few potential candidate models and make the transfer more informative than using a single model. 

However, the models discarded by this selection process may still contain useful knowledge. By analogy with multimodal problems, one may consider each independent model (specifically the last hidden representation of the neural networks) as a modality for the new task. Some modalities taken in isolation can yield bad results while they carry useful information complementary to other modalities. On the contrary, two well-performing modalities can be redundant and combining them brings less improvement. 

Extending this reasoning to $M$ modalities can be done by learning a common representation embracing all the modalities. However, this is not a trivial problem. For example, naively concatenating the local embeddings provided by each model does indeed produce a common embedding, but does not work well when dealing with many tasks/modalities, as observed by Zamir \etal~\cite{zamir2018taskonomy}. 

We can formulate this multi-source transfer learning as follows. We define knowledge items as (domain, task) pairs, which can be associated with a model trained on this domain to fulfill this task. Supposing having $M$ of such source knowledge, how to group them into a unique and general model of knowledge, performing well both on source knowledge and on novel target knowledge? 

By addressing such a goal, this paper makes three main  contributions:
\textbf{(a)} the definition of a  carefully designed ensemble of source and target knowledge related to facial analysis. Facial analysis is a topic of broad interest having received a lot of attention from the community, and for which large sets of knowledge (\ie pre-trained models) are available. Moreover, a general knowledge in such a field would be of practical interest, as the number of new face-related tasks and domains grows exponentially. \textbf{(b)} a simple yet efficient methodology to project the source embeddings into a unique one, which is accurate both on the source and the target knowledge.
\textbf{(c)} A distillation process is introduced to transfer the learned general knowledge into one lightweight convolutional neural network. This simple model outperformed its teacher, being on par with state-of-the-art models specifically built for solving specific tasks on specific domains. Moreover, it fits real-world application requirements, with 2 million parameters.

\section{Related Work}
The proposed work is related to several fields including information compression, transfer learning or distillation.

\vspace{-1em}\paragraph{Information Fusion and Multi-Task Learning}
Combining the knowledge from several existing models into a single representation can be seen as a multimodal fusion, where model features are the input modalities. Classical methods exist such as Principal Components Analysis~\cite{joliffe1992principal} or Canonical Correlation Analysis~\cite{hardoon2004canonical}. The recent trend is often focusing on equivalent neuronal methods such as the one Ngiam \etal~\cite{ngiam2011multimodal} proposed, using multimodal autoencoders to learn a representation able to reconstruct all the modalities. The literature on deep multimodal fusion can  be divided in i) multimodal architectures~\cite{snoek2005early, neverova2016moddrop, perez2019mfas} focusing on where to fuse the information in the network, ii) representations based on constraints~\cite{ andrew2013deep,chandar2016correlational,shahroudy2018deep} building on the relations between the modalities (\eg correlations). More details about this field can be found in these two surveys~\cite{atrey2010multimodal,baltruvsaitis2019multimodal}.

Information fusion and multitask learning approaches are often related, as in the multitask autoencoder proposed by Ghifary \etal~\cite{ghifary2015domain} allowing  to improve domain generalization. The recent work by Ruder \etal~\cite{ruder122019latent} on multitasking architecture search allows to learn latent architectures for multitasking problems. Multitask approaches for face analysis have been proven to be as efficient or even better than single-task learning~\cite{chang2019deep}.
In the context of this paper, we can also see the fusion of diverse existing models both as a multimodal problem (each model is a modality) and as a multitask problem (the final student model has to be efficient on all tasks).

\vspace{-1em}\paragraph{Transfer Learning} 
As the goal of the paper is to transfer  knowledge to novel tasks, it can be related to transfer learning. We share the same motivations as Taskonomy proposed by Zamir \etal~\cite{zamir2018taskonomy}, helping to select which combination of existing models to use when tackling a new task. Nevertheless in our case, because the existing models are already partially related to the target task, there is a benefit to select them all and keep what is useful in each model. Geyer \etal~\cite{geyer2018transfer} propose to merge two pre-trained models before transferring them, based on incremental moment matching~\cite{lee2017overcoming}. Chen \etal~\cite{chen2018coupled} designed a coupled end-to-end transfer learning, distilling the knowledge of one source model into the target model, while selective adversarial networks are proposed in~\cite{cao2018partial} to select positive transfers and discard negative ones in the particular case where the target label space is a subspace of the source label space. Finally, regarding domain generalization, Mancini \etal~\cite{mancini2018best} propose to fuse the outputs of domain-specific neural networks, after predicting the domain of target samples.

\vspace{-1em}\paragraph{Self-supervised Learning}
Our work uses a large dataset to learn an autoencoder in an unsupervised fashion, and, by this means, exhibits the relations between the given models. Therefore, training the Teacher can be seen as a self-supervised learning process, which is a widely explored topic~\cite{zhang2016colorful,zhang2017split,noroozi2017representation}. The goal of self-supervised learning is to design an efficient and cost-less proxy task helping to solve the target task for which we don't have enough annotations. Doersch \etal~\cite{doersch2017multi} showed the benefit of using several proxy tasks in self-supervised learning, which is basically what we are doing with the six different embeddings to reconstruct. The difference lies in the very definition of the proxy task. Ours are coming from previously learned knowledge. In traditional self-supervised, the tasks are low-level objectives such as solving a jigsaw puzzle~\cite{noroozi2016unsupervised} or evaluating the rotation of an image~\cite{gidaris2018unsupervised}. Radenovic \etal~\cite{radenovic2018deep} used a proxy task closer to ours, using state-of-the-art models to extract edges of images as labels for their visual model.

Other methods falling in this unsupervised learning category can also be linked to ours, such as the deep clustering approach~\cite{caron2018deep} where a convolutional neural network if trained using the output of a k-means clustering algorithm. 

\paragraph{Model Compression and Knowledge Distillation}
A last important aspect of our method is to distill~\cite{HintonVD15} the learned knowledge into a single lightweight model. In this spirit, Romero \etal proposed a deep and thin student named Fitnet~\cite{romero2014fitnets}. It learns from both last layer and hidden layers of the teacher and outperforms it. Aiming to ensure privacy of the learning dataset, Papernot \etal~\cite{papernot2016semi} proposed to use several teachers in a semi-supervised fashion.

With a goal close to ours, Chebotar \etal~\cite{chebotar2016distilling} used as a teacher a weighted average prediction of an ensemble of neural networks.
More recently the idea of data distillation was developed by~\cite{radosavovic2018data}, applying the same teacher model to diverse transformations of the input image and taking the average prediction as a label for the student. Li \etal~\cite{li2019delta} use feature map  attention to regularize the learning of the student.
Finally, multimodal distillation bears similarities with our approach, using some well-known modalities of a given input to master other views with fewer annotations. For instance, the SoundNet model~\cite{aytar2016soundnet}  learns an audio model from the labels yield by a visual model, while Xu \etal~\cite{xu2018pad} proposed to fuse the predictions coming from diverse modalities to improve the quality of the final main task of the student.

\section{Methodology}
\label{methods}
\begin{figure*}[ht]
    \centering
    \includegraphics[width=0.9\textwidth]{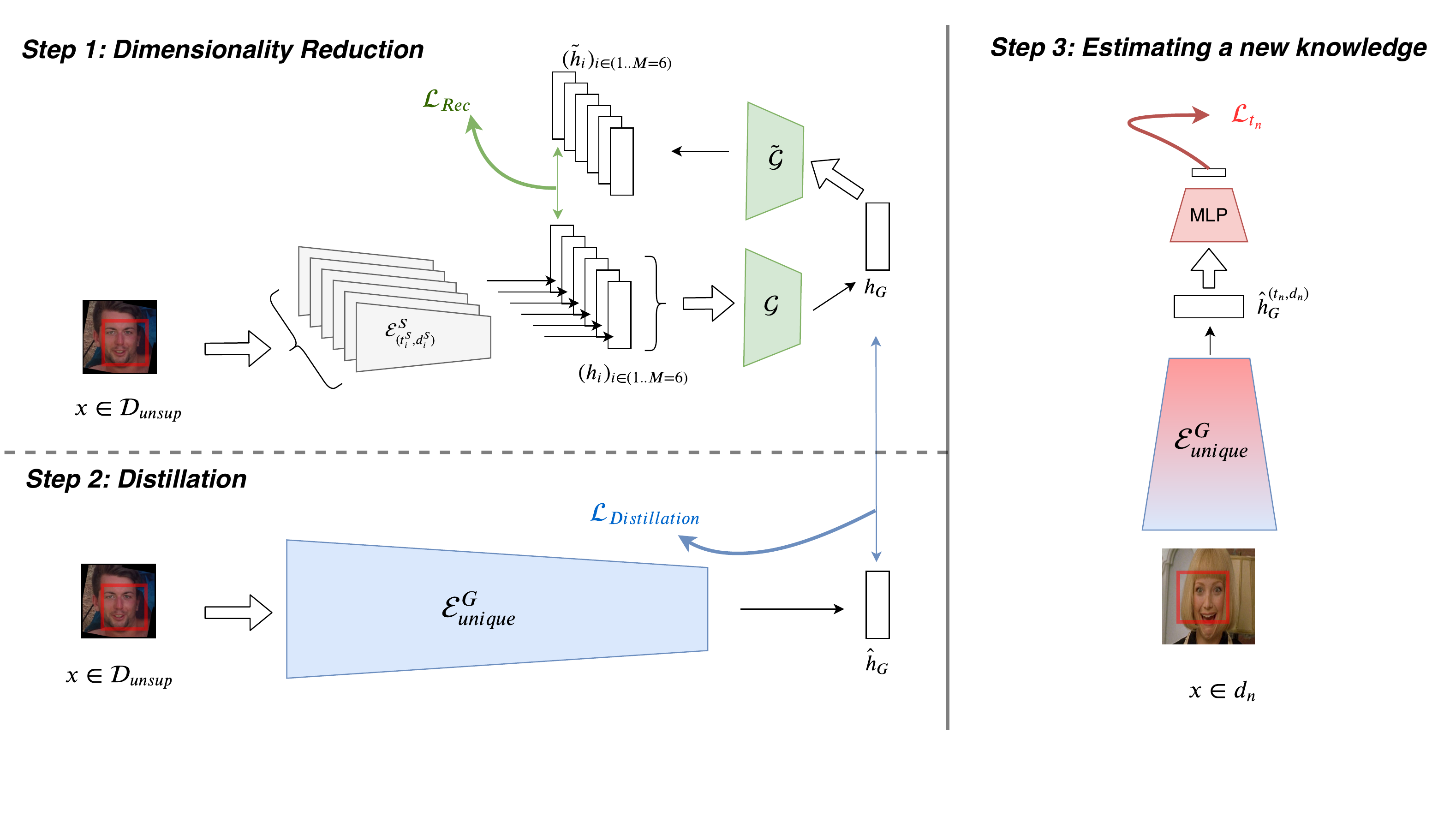}
    \vspace{-1cm}
    \caption{Overview of the method. \emph{Step 1} is the training of $\mathcal{G}$ on $\mathcal{D}_{unsup}$, to obtain an encoder $\mathcal{E}^G$ of $x$ in a compact embedding $h_G$, gathering all sources of knowledge (represented by the extracted $h_j$). \emph{Step 2} aims at reducing the number of parameters used to get $h_G$ by distilling the knowledge of 130 millions parameters into a much simpler encoder $\mathcal{E}^G_{unique}$, leading to $\hat{h}_G$. In \emph{step 3} this lightweight encoder $\mathcal{E}^G_{unique}$ can then be plugged with a MLP $\mathcal{C}_{(t_n,d_n)}$ and all parameters adapted to learn new knowledge $\mathcal{K}_{(t_n,d_n)}$.}
    \label{fig:overview}
\end{figure*}

To the best of our knowledge, transferring multi-source knowledge across both tasks and domains has not been addressed yet in the literature. Therefore, we will first study how to formally and practically formalize such a problem. We will then detail the two main steps composing our approach: dimensionality reduction and distillation.

\subsection{Multi-source  Multi-domain Transfer Learning}
\paragraph{General formulation}
Let's first define the concept of {\em knowledge} as the abstract ability to perfectly solve a given task $t$ on a given domain $d$. We limit ourselves to the family of tasks including classification / regression with machine learning techniques. The knowledge extracted when solving this problem (\eg through the training of a deep neural network) can be denoted as $\mathcal{K}_{(t,d)} = (\mathcal{E}_{(t,d)}, \mathcal{C}_{(t,d)})$, where
\begin{itemize}
    \item $\mathcal{E}_{(t,d)}$ is a function able to map each element $x$ of domain $d$ into a common embedding $h$
    \item $\mathcal{C}_{(t,d)}$ is a function able to map each embedding $h$ to the expected output $y$
\end{itemize}

When dealing with a new target problem defined by a task $t'$ on a domain $d'$, $\mathcal{K}_{(t,d')}$ or $\mathcal{K}_{(t',d)}$ can be used to initialize a proposal for  $\mathcal{K}_{(t',d')}$, that can be further adapted to fit the task/domain. 
\textit{Transfer learning} consists in reusing (or fine-tuning) $\mathcal{E}_{(t,d)}$ and learning a new specific $\mathcal{C}_{(t',d)}$. While \textit{domain adaptation} leads to learn $\mathcal{E}_{(t,d+d')}$ and $\mathcal{C}_{(t,d+d')}$.

The general problem we tackle is the one of adapting transfer learning and domain adaptation to the case where we have not only one already solved problem, but a set $\mathcal{S}$ of solved problems. 

Following transfer learning and domain transfer, our approach is based on defining an operator $\mathcal{E}^G = \mathcal{G}({\mathcal{E}_{(t_i,d_i)}, i \in \mathcal{S}})$  where $\mathcal{G}$ is a combination function using all the $h_i$ embedding (coming from the $\mathcal{K}^S_{(t^S_i,d^S_i)}$). It aims at gathering all knowledge information regarding the $t_i$ and the $d_i$ into one embedding $h_G$.

One possible solution is to concatenate all the outputs of ${\mathcal{E}_{(t_i,d_i)}, i \in \mathcal{S}}$ to generate $h_G$. But as observed in Taskonomy~\cite{zamir2018taskonomy}, it generally suffers from lack of generalization.

Recent approaches such as Taskonomy~\cite{zamir2018taskonomy,achille2019task2vec} have addressed this question by selecting the $\mathcal{K}_{(t_i,d_j)}$ where $t_i$ and $t_j$ are most correlated. 

In our approach (see Section~\ref{teacher}),  $\mathcal{G}$ is a neuronal encoder allowing to estimate a representation $h_G$ of reduced dimensionality, still approximating well the $h_i$ and potentially leading to better generalizations by removing biases due to over-complexity. 

Nevertheless, $\mathcal{E}^G$ is then composed of all $\mathcal{E}_{(t^S_i, d^S_i)}$ and of $\mathcal{G}$ and thus mapping a given $x$ to $h_G$ has a high computational cost. In Section~\ref{distill} we propose the use of {\em distillation} to transform $\mathcal{E}^G$ into a unique and lightweight model $\mathcal{E}^{G}_{unique}$, directly projecting $x$ to $\hat{h}_G$ and therefore allowing easier transfer when using it to get a new knowledge.

\paragraph{Source and target knowledge for facial analysis}
As building a general knowledge on all tasks and domains existing would not be feasible, we propose to validate our approach experimentally by carefully designing a set $\mathcal{S}$ of Source-Knowledge and a set $\mathcal{T}$ of Target-Knowledge  to cover a facial analysis general knowledge, including the 15 different knowledge described in Table~\ref{tab:knowledge}. 

$\mathcal{S}$ gathers M=6 source knowledge, which leads to a domain $\bigcup\limits_{i}d^S_i$ of around 22 million faces with M=6 different tasks contains in around 106 million parameters. To train $\mathcal{G}$ (in an unsupervised fashion), instead of using $\bigcup\limits_{i}d^S_i$,  we will use a domain $\mathcal{D}_{unsup}$ of 4.12 millions of faces (discarding annotations), extracted from VGGFace2~\cite{cao2018vggface2} (3.14 million), EmotioNet~\cite{fabian2016emotionet} (0.72 million) and IMDb-WIKI~\cite{imdbwiki} (0.26 million).

$\mathcal{T}$ contains 9 various knowledge on facial analysis. Note that two tasks with the same name (for instance Expression Classification) may still be different (for instance facial expression is subjective and depends on its annotators~\cite{zeng2018facial}).

\begin{table}[ht]
\centering
\begin{tabular}{lll}
\hline
$\mathcal{K}_{(t_i,d_i)}$                     & $t_i$                                     & $d_i$ size \\ \hline
\textbf{Expr-AffectNet} & Emotion Classif.~\cite{vielzeuf2018occam}             & 0.3M~\cite{mollahosseini2017affectnet}      \\
Expr-RAF                & Emotion Classif.~\cite{vielzeuf2018occam}             & 15,339~\cite{raf}      \\
Expr-SFEW               & Emotion Classif.~\cite{vielzeuf2018occam}           & 1,766~\cite{sfew}       \\
\textbf{Identity-MS}          & Identity Matching~\cite{schroff2015facenet}              & 6.5M~\cite{guo2016ms}    \\
Identity-LFW                  & Identity Matching~\cite{schroff2015facenet}                        & 13,000~\cite{lfw}      \\
\textbf{Gender-IMDb}          & Gender Prediction~\cite{antipov2017effective}                         & 0.5M~\cite{imdbwiki}    \\
Gender-UTK                    & Gender Prediction~\cite{das2018mitigating}                         & 20,000~\cite{celebA}      \\
\textbf{Attrib-CelebA}        & Attrib. Detection~\cite{cao2018partially}                   & 0.2M~\cite{celebA}     \\
\textbf{AgeR-IMDb}          & Age Regression~\cite{imdbwiki}                            & 0.5M~\cite{imdbwiki}     \\
AgeR-FG                     & Age Regression~\cite{antipov2017effective}                           & 1,000~\cite{fgnet}       \\
AgeR-UTK                    & Age Regression~\cite{das2018mitigating}                            & 20,000~\cite{utk}       \\
AgeC-UTK                & Age Classif.~\cite{das2018mitigating}         & 20,000~\cite{utk}       \\
Ethnic-UTK                      & Ethnicity Classif.~\cite{das2018mitigating}      & 20,000~\cite{utk}       \\
Pain-UNBC                     & Pain Estimation~\cite{zhang2018bilateral} & 48,000~\cite{pain}      \\
\textbf{Object-ImageNet}      & Object Classif.~\cite{deng2009imagenet}      & 14M~\cite{deng2009imagenet}  \\ \hline
\end{tabular}
\caption{The M=6 source and 9 target knowledge (source is in bold). The task, as well as the used pre-trained model for source knowledge, are described in the associated papers. Information about the used domain can be found in the second citation on each row. Moreover, more details about all the selected knowledge will be provided in supplementary materials.}
\label{tab:knowledge}
\end{table}

\subsection{Dimensionality Reduction for a More General Knowledge}
\label{teacher}
We explain in the previous subsection that we dispose of M source knowledge encoders $\mathcal{E}_{(t^S_i, d^S_i)}$, allowing to extract M $h_i$ embedding from a given face $x$. We study here the operator $\mathcal{G}$ allowing to combine these embedding into a compact and general embedding $h_G$.

\paragraph{Motivations}
Defining $\mathcal{G}$ as a basic concatenation function implies a very large and redundant $h_G$. Therefore there is a meaning in reducing the dimension of $h_G$, leading to discard some redundancies and exploit the complementarity between the $h_j$. Dimensionality reduction is a well-explored topic, from linear projections such as PCA~\cite{joliffe1992principal} to more complex non-linear approaches such as manifold learning~\cite{huo2007survey}. The major drawback of a linear approach is that the representation is projected into a plane and therefore can miss the real shape of the data. Thus, learning the manifold represented by $h_G$ with a (non-linear) neural network~\cite{li2018measure} makes sense. 

\paragraph{Adopted Approach}
As we want to reduce the dimensionality without supervision, we propose to train $\mathcal{G}$ with an auto-encoding objective $\mathcal{L}_{Rec}$, as shown in Step 1 of Figure~\ref{fig:overview}. In other words, we are building $h_G$ as an answer to all source tasks (represented by the $h_j$) but on the $\mathcal{D}_{unsup}$ domain.
For that, we optimize the parameters of both $\mathcal{G}$ (as an encoder) and of $\tilde{\mathcal{G}}$ (as a decoder reconstructing the $h_i$):
\begin{align}
    h_G = \mathcal{G} ((h_i)_{i=1..M})\\
    (\tilde{h}_i)_{i=1..M} = \tilde{\mathcal{G}}(h_G) \\
    \mathcal{L}_{Rec} = \Sigma_{m=1}^M || \hat{h}_i - h_i ||^2
\end{align}
Note that we choose $\tilde{\mathcal{G}}$ to have a symmetric architecture to $\mathcal{G}$.
We will further discuss the architecture choice of an autoencoder in the experiments section, by also experimenting with PCA, regular autoencoders~\cite{ballard1987modular}, variational autoencoders~\cite{kingma2013auto} and denoising autoencoders~\cite{vincent2010stacked}. 

Another important choice is the choice of the dimensionality of $h_G$, as it drives the knowledge compression process. As we are aiming at creating a unique general knowledge, we consider as a rule of thumb to fix it as the average dimensionality of the $h_i$. We will conduct a empirical study in subsection~\ref{rep_influence} on the impact of this dimension on the quality of $h_G$ and check that there is no optimal dimension, only extreme values (very low-dimensional or high-dimensional) clearly degrading our approach.

\subsection{Real-world Transfer Learning by Distillation}
\label{distill}
As mentioned in the problem definition, obtaining $h_G$ by step 1 implies using $\mathcal{E}^G$, composed of M=6 different pre-trained models combined to $\mathcal{G}$. It leads to a huge number of parameters (130M) and greatly limits the possibility of domain adaptation when dealing with the target knowledge, as we can't adapt such a large number of parameters on a new domain. A natural way to solve this problem would then to compress the so-obtained model. Model compression can be addressed with several approaches as shown in the recent literature~\cite{cheng2017survey}, mainly gathered into four categories: parameters pruning, low-rank factorization, compact convolutional filters and knowledge distillation. 

Our current model is composed of M=6 different specific branches and we do not only want to reduce the number of parameters: we want to achieve a unique encoder. It is the promise of the distillation approach, allowing to train a new \emph{Student} model, supervised by the previous big \emph{Teacher} model~\cite{HintonVD15}. Thus, we will not only make $\mathcal{E}^G$ lighter but also transform it to a conventional estimator $\mathcal{E}^G_{unique}$, such as a classic convolutional neural network architecture, on which well-known methods such as data augmentation can be easily applied. Moreover previous works~\cite{chu2016best} have shown that fine-tuning all the parameters of a model may lead to better domain adaptation and improve the quality of the transfer.

\paragraph{Distillation}
To achieve distillation, we still are training our new $\mathcal{E}^G_{unique}$ on $\mathcal{D}_{unsup}$, as illustrated in Figure~\ref{fig:overview}.
We consider $\mathcal{E}^G_{unique}$ as a neural network of arbitrary architecture (we choose a ResNet-18~\cite{he2016deep} for all experiments), taking the face image $x$ and directly projecting it into a representation $\hat{h}_G$.

The training is down by minimizing $\mathcal{L}_{Distillation}$
\begin{equation}
   \mathcal{L}_{Distillation} = D_c(h_G , \hat{h}_G)
\end{equation}
where $D_c$ is the cosine metric (\ie $D_c(a,b) = \frac{a^\top b}{\|a\|\cdot\|b\|}$).

\paragraph{Using $\mathcal{E}^G_{unique}$ to estimate knowledge from $\mathcal{T}$}
Finally when $\mathcal{E}^G_{unique}$ has been trained, the last step is straight-forward and only consists in adding a Multi-Layer-Perceptron on top of it and train all parameters on to estimate the new target knowledge.

\subsection{About Using 2 Step Training}
The two previous steps may be seen as training two parts of the same model and may be done at the same time, by training $\mathcal{E}^G_{MT} =  \tilde{\mathcal{G}} \circ \mathcal{E}^G_{unique}$ to directly fit the $h_i$. This approach may be considered as a \textbf{weakly supervised multi-task (MT) learning}. In practice we observe that it does not converge as well as our two-step approach, which is disentangling the processing relative to the tasks (first step) and to the domains (second step). It also is in line with several progressive approaches observed in the literature~\cite{karras2017progressive}, achieving better model convergence by progressively training different parts of the model.

\section{Experiments}
\label{experiments}
This section validates our approach, first by detailing results of $\mathcal{E}^G_{unique}$ on the different source and target knowledge, compared to state-of-the-art dedicated approaches, then by running an ablation study on the different steps of our method and on the benefit to decompose the learning process into these steps.

\subsection{Implementation details}
We pre-process all faces from all domains with same operations. We use a private face detector to first detect and loosely crop the detected faces. If more than one face is detected, we select the closest to the image center. A landmark-based aligner is then applied on the detected faces, which are finally resized to $300\times300\times3$ pixels. 
Table~\ref{detection} reports the number of images where a face was detected for each domain (first column). When a face is not detected (second column), we will consider during evaluation of our models (a) that the prediction is not correct when addressing classification problems (b) and that the prediction is the average between minimum and maximum values when dealing with regression problems. Note that the rest of the paper is following this rule.

\begin{table}[tb]
\centering
\begin{tabular}{lll}
\hline
Domain      & \# Detected Faces  & \# Undetected Faces \\ \hline
CelebA       & 202442 & 177        \\
UTKFace      & 24018  & 98          \\
FG-NET       & 1002    & 0            \\
SFEW         & 1732   & 37          \\
RAF          & 15330  & 9        \\
ShoulderPain & 48391  & 0            \\
LFW          & 13233  & 0            \\
$\mathcal{D}_{unsup}$ & 4.12 M  & 0.02M          \\ \hline
\end{tabular}
\caption{Number of detected/undetected faces for each domain.}
\label{detection}
\end{table}

As we use different tasks and domains, the evaluation protocols are changing for each knowledge. We follow the exact evaluation protocol used by the state-of-the-art dedicated approaches we are comparing to. 
\subsection{Experimental Validation}
We first propose to evaluate our final approach $\hat{h}^{(t,d)}_G$ on target knowledge. Looking at the two last columns of Table~\ref{tab:cnn_results}, we observe that our approach adapt well to all target knowledge, even outperforming dedicated state-of-the-art approaches on AgeC-UTK, Expr-SFEW, Expr-RAF, Ethnic-UTK, AgeR-UTK. Note that the model for $\hat{h}^{(t,d)}_G$ counts only 2.2 millions parameters, which is almost always far less than the parameters used by other approaches, helping to better generalize and better fits real-world applications constraints.
\paragraph{Target Knowledge}
\begin{table*}[ht]
\centering
\begin{tabular}{ll||ll|lllll|l}
\hline
Dataset       & Metric     & CNN & P-CNN    & $\mathcal{E}^G$  & \multicolumn{2}{c}{$\mathcal{E}^G_{MT}$} & \multicolumn{2}{c}{$\mathcal{E}^G_{unique}$}          & State-of-the-art (\# parameters)                                                       \\ \hline

              &            &    &    & $h_{G}$ & $\hat{h}_{MT}$        &$\hat{h}^{(t,d)}_{MT} $         & $\hat{h}_G$     &$\hat{h}^{(t,d)}_G$                      &                                                                                        \\\cmidrule(lr){5-5}\cmidrule(lr){6-7} \cmidrule(lr){8-9} 
AgeC-UTK      & Acc.    & 57.80 &64.30  & 68.80  & 67.90      & 68.40       & 68.80   & \textbf{70.40}            & 70.10~\cite{das2018mitigating}(5 M)                                                     \\
Gender-UTK    & Acc.     & 90.00 &96.50 & 97.20     & 93.15     & 94.4       & 97.58  & 97.90                   & \textbf{98.23~\cite{das2018mitigating}} (5 M)                                          \\
Expr-SFEW     & Acc.     & 22.00 &53.60  & 52.10      & 50.20      & 52.20       & 54.00     & \textbf{57.2}           & \textbf{55.40-58.14~\cite{vielzeuf2018occam}~\cite{acharya2018covariance}} (1.7 M/5 M) \\
Expr-RAF      & Acc.     & 69.00 &85.70  & 86.51     & 82.40      & 87.20       & 87.30  & \textbf{89.3}         & 86.77~\cite{zeng2018facial} (35 M)                                                     \\
Ethnic-UTK      & Acc.     & 62.20 &84.90   & 89.20    & 81.20     & 82.20       & 88.2   & \textbf{91.20}         & 90.10~\cite{das2018mitigating} (5 M)                                                    \\
Identity-LFW  & Acc.      & (98.40) &(99.65) & 99.10  & 94.27     & (99.1)     & 98.92  & (99.42)                 & \textbf{99.65-99.87}~\cite{schroff2015facenet}(25 M)  
\\
\hdashline
AgeR-UTK      & MAE        & 6.38 &4.70  & 4.39    & 4.70       & 4.42       & 4.24   & \textbf{4.05}         & 5.39~\cite{cao2019consistent}(21.8 M) 
\\ 
Pain-UNBC     & MAE        & 0.89 &0.56 & 0.54    & 0.64      & 0.72       & 0.56   & 0.52           & \textbf{0.51~\cite{zhang2018bilateral}} (0.0001 M)                                     \\
AgeR-FG       & MAE        & 11.10 &3.10 & \textbf{2.85}  & 3.95      & 3.12       & 2.95   & 3.05            & \textbf{2.81-3.00}~\cite{rothe2018deep}~\cite{antipov2017effective} (25 M)              \\
\hdashline
Attrib-CelebA & ER & 8.60 &8.04  & 7.70    & 8.12      & 8.04       & 7.81   & 7.67                   & \textbf{7.02~\cite{cao2018partially}}(16 M)                                            \\           
\hline
\end{tabular}
\caption{Performance of diverse end-to-end approach evaluated on the target knowledge. All approaches are described in the methods or detailed in the text. Note that for the specific case of LFW, two types of results are reported. For result between parenthesis, before evaluation on LFW, the network is first trained (or fine-tuned) on a subset of 100,000 faces from VGGFace2 to predict Identity. The other LFW results are obtained directly from embedding $h$. Acc. is Accuracy and ER is the Average Error Rate.}
\label{tab:cnn_results}
\end{table*}
Moreover, this level of performance with a lightweight model is difficult to achieve, as shown by the results of the CNN baseline, which has the same architecture than $\hat{\mathcal{K}}^G_{unique}$ but is trained from scratch on the domain of the target knowledge. Indeed, this CNN is not always converging (\eg on AgeR-FG with only 1002 images) and there is a huge gap between its scores and our approach, highlighting the clear benefit of transferring knowledge. A second stronger baseline called P-CNN is basically the classic way for transfer learning: it consists in the use of a pre-trained CNN, chosen among the source encoders as the one transferring the best for the target task and then fine-tuned on the target task for all its parameters. See also Table~\ref{tab:ae_results} for the scores obtained by selecting the best source encoder but without fine-tuning all parameters. Note that fine-tuning all parameters does not lead always to an improvement.

\paragraph{Source Knowledge}
We also analyze the performance of $\hat{h}^{(t,d)}_G$ on the source knowledge and compare it to the original pre-trained models used as source knowledge estimators in Table~\ref{tab:origin}.
For Expression-AffectNet, Attrib-CelebA and Gender-UTK we observe a clear improvement, while for Age-FG and for Identity-LFW the performance is slightly degraded.
For Age-FG it may be explained by performance saturation, as the domain is very small and the human performance is around 4.6 in MAE~\cite{han2013age}. For Identity-LFW, the loss can come from the difference in representation size: the original model representation has 2048 features dedicated to Identity Matching, while our model counts only 1024 features. Augmenting the dimension of $\hat{h}_g$ may lead to more comparable performance and will be discussed in subsection~\ref{rep_influence}.
Finally, to avoid a long training time we do not compare the models on ImageNet but on the smaller dataset TinyImageNet~\cite{yao2015tiny}. The obtained accuracy (on the validation set) clearly underlines the importance in the choice of $\mathcal{D}_{unsup}$: the unsupervised training has been done only on face images and thus is not beneficial for tasks such as ImageNet classification. Despite this loss in performance, note that when training a ResNet-18 from scratch on TinyImageNet, we achieve only 52\% accuracy.

\subsection{Ablation study}
\paragraph{Contribution of $\mathcal{G}$}
As described in Section~\ref{methods},  $\mathcal{G}$ takes as input the concatenation of the 6 embeddings $(h_i)$, one per task (of size 5488 in our case). Then, the goal is to generate a compact representation $h_G$ (which size is fixed to 1024 in all the experiments) of the M=6 embeddings $(h_i)$. As this dimension reduction step is a crucial operation of our approach, we propose to study the impact of different variations on this step.

\begin{table}[bt]
\centering
\begin{tabular}{lllll}
\hline
Knowledge& Metric  & Original & $\mathcal{E}^G$ &   $\mathcal{E}^G_{unique}$             \\ \hline
Expr-AffectNet&{\small Accuracy}         & 63.5                         & 64.0                        & \textbf{64.4} \\
Identity-LFW& {\small Accuracy}              & \textbf{99.65}               & 99.1                        & 99.42         \\
Gender-UTK& {\small Accuracy}            & 96.5                         & 97.2                        & \textbf{97.9} \\
{\small Object-TinyImageNet}& {\small Accuracy}                    & \textbf{76.2}                & 71.8                        & 56.5  \\
\hdashline
AgeR-FG      & MAE               & \textbf{2.85}                        & \textbf{2.85}                        & 3.05          \\
\hdashline
Attrib-CelebA      & {\small Error rate}         & 8.03                         & 7.7                         & \textbf{7.67} \\
\hline
\end{tabular}
\caption{Performance of the original models $\mathcal{E}^S_{(t^S_j,d^S_j)}$ , of $\mathcal{E}^G$ and of $\mathcal{E}^G_{unique}$ on the original tasks.}
\label{tab:origin}
\end{table}

Therefore, we first evaluate the ability of $\mathcal{G}$ to reconstruct the original $h_i$ features from $h_G$. We report in Table~\ref{table:recon} the normalized Root Mean Squared Error between each $h_i$ and $\tilde{h}_i$ with diverse variations on $\mathcal{G}$. PCA stands for Principal Components Analysis, AE for standard Autoencoder, VAE for Variational Autoencoder and DAE for Denoising Autoencoder. Note that AE, VAE and DAE have the same number of parameters (the encoder $\mathcal{G}$ has 3 fully connected layers: 5488$\times$3136, 3136$\times$1792 and 1792$\times$1024 and the decoder $\tilde{\mathcal{G}}$ is the symmetrical). 
The linear PCA baseline is easily outperformed by other approaches. The two best performing are the AE and the DAE. Surprisingly, the DAE does not generalize better than the AE, and is performing better only for the reconstruction of the ImageNet embedding, which is the less relevant in a context of face inputs. Still note that the gap of performance is not very important and all non-linear methods allows a fairly decent reconstruction of the embeddings.

\paragraph{Contribution of the distillation: $\mathcal{E}^G_{MT}$ versus $\mathcal{E}^G_{unique}$}
The last two columns of the Table~\ref{table:recon} allow to evaluate $\mathcal{E}^G_{MT}$ and $\mathcal{E}^G_{unique}$, by their ability to reconstruct the $h_i$ from the embedding they produced. To achieve such a reconstruction, we extracted the $\hat{h}_{MT}$ and $\hat{h}_G$ from all element of $D_{unsup}$ and then trained a decoder (with a similar architecture to $\tilde{\mathcal{G}}$) to reconstruct the $h_i$.

\begin{table}[tb]
\centering
\begin{tabular}{lllll|ll}
\hline
$h_i$ & PCA  & AE            & VAE  & DAE           & $\mathcal{E}^G_{MT}$   & $\mathcal{E}^G_{unique}$ \\ \hline
Expr     & 0.28 & \textbf{0.23} & 0.26 & 0.26          & 0.55 & 0.39          \\
Identity & 0.25 & \textbf{0.22} & 0.23 & 0.26          & 1.12 & 0.35          \\
Object   & 0.48 & 0.26          & 0.30  & \textbf{0.25} & 0.67 & 0.44          \\
Age      & 0.23 & \textbf{0.19} & 0.22 & \textbf{0.19} & 0.77 & 0.38          \\
Attrib   & 0.33 & \textbf{0.27} & 0.31 & 0.29          & 0.97 & 0.35          \\
Gender   & 0.25 & \textbf{0.25} & 0.27 & 0.27          & 1.22 & 0.30           \\ \hline
Average  & 0.31 & \textbf{0.24} & 0.27 & 0.25          & 0.92 & 0.37          \\ \hline
\end{tabular}
\caption{Reconstruction normalized RMSE obtained on $\mathcal{D}_{unsup}$ test set by considered methods for each source knowledge embedding $h_i$ and on average.}
\label{table:recon}
\end{table}

We can observe that the reconstruction error of $\mathcal{E}^G_{MT}$ is sometimes very high (\eg for the identity embedding, explaining the low results of $\hat{h}_{MT}$ when use for Identity-LFW in Table~\ref{tab:cnn_results}). Note that simply generating random embeddings (according to a uniform distribution in the range of the target embeddings) allows to achieve a normalized RMSE of 1.4. Therefore a score of 1.22 is almost random.

Without surprise, the reconstruction error of $\mathcal{E}^G_{unique}$ is higher than the one obtained by $\mathcal{G}$, which can be explained by both the distillation approximation between $h_G$ and $\hat{h}_G$ and the lower capacity of our small model. Nevertheless, a low reconstruction error is not a guarantee of better performance when trying to apply the $h_G$ representation in other tasks and domains. Thus, we propose to evaluate the very transfer learning operation using $\mathcal{E}^G$ instead of $\mathcal{E}^G_{unique}$. The first column of Table~\ref{tab:cnn_results} reports the results obtained by $\mathcal{E}^G$ on the target knowledge, while Table~\ref{tab:origin} provides the results on the source knowledge. On almost all knowledge, $\mathcal{E}^G$ is outperformed by $\mathcal{E}^G_{unique}$ by a significant margin.

Yet, if we try to understand from where the distillation improvement is coming, we can observe that only using a MLP on top of $\hat{h}_G$ (without fine-tuning all parameters) does not bring improvements compared with using a similar MLP on top of $h_{G}$, some small changes in term of performance being observed. Thus, the distillation error has a limited impact when dealing with transfer. The bigger improvement is observed when using $\hat{h}^{(t,d)}_G$, meaning that we fine-tune all parameters of $\mathcal{E}^G_{unique}$ on the new knowledge. It illustrates that the main contribution of the distillation lies in the reduced size of the encoder, giving it the ability to more easily adapt to new tasks and domains.

\paragraph{Concatenation, reduction with $\mathcal{G}$ or selection ?}
If we come back to the operator $\mathcal{G}$, we discussed in section~\ref{methods} the good reasons to reduce dimensionality of the $h_i$. We propose to empirically compare this choice to other methods and validate the intuition that selection may discard useful information.
Thus, for each target knowledge, we propose to evaluate $\mathcal{G}$ and several alternatives summarized in Table~\ref{tab:ae_results}.  
We first evaluate a  Multi Layer Perceptron taking as input the concatenation (Concat of all $h_i$) and having $\mathcal{G}$ architecture. The MLP parameters are trained from scratch on the target knowledge.
Our $\mathcal{G}$ is then evaluated by training a small MLP on top of $h_G$ (extracted from all images of the target domain).
We also compare the benefits brought by using a non-linear $\mathcal{G}$ by also reporting the Princpal Components Analysis (PCA) results.

Finally, another concurrent approach consists in selecting the embedding (resp. the combination of embedding) yielding the Best Transfer (BT) (resp. the Best Combination Transfer (BCT)). Several works ~\cite{zamir2018taskonomy,li2019task} proposed a method to automatically select such a best combination of embeddings. We choose here to reproduce this approach in a naive way, by brute force testing all the possible single transfers (BT) or combinations (by concatenation) of transfers (BCT) and reporting the best found results on each target kowledge in the two last columns of Table~\ref{tab:ae_results}.

\begin{table}[tb]
\centering
\begin{tabular}{llllll}
\hline
Knowledge       & \multicolumn{3}{c}{$\mathcal{G}$}     & \multicolumn{2}{c}{Selection} \\ \hline
              & Concat & PCA  & AE             & BT        & BCT               \\ \cmidrule(lr){2-4} \cmidrule(lr){5-6}
AgeC-UTK      & 65.50   & 65.00   & \textbf{68.80}  & 63.2      & 67.20              \\
Gender-UTK    & 96.92  & 96.7 & \textbf{97.20}  & 96.50      & 97.10              \\
Expr-SFEW     & 45.70   & 32   & 52.10           & 52.20      & \textbf{53.1}     \\
Expr-RAF      & 84.89  & 81.9 & \textbf{86.51} & 85.48     & 85.74             \\
Ethnic-UTK      & 86.65  & 62.5 & \textbf{89.20}  & 83.40      & 86.20              \\
Identity-LFW  & 88.10  & 96.8 & 99.10           & 99.65     & \textbf{99.7}     \\
\hdashline
AgeR-UTK      & 4.45   & 4.68 & \textbf{4.39}  & 4.70       & 4.54              \\
Pain-UNBC     & 0.69   & 0.6  & 0.54           & 0.53      & \textbf{0.51}     \\
AgeR-FG       & 3.22   & 3.18 & \textbf{2.85}  & \textbf{2.85}      & \textbf{2.85}            \\
\hdashline
Attrib-CelebA & 8.07   & 8.03 & \textbf{7.70}   & 8.03      & 7.85              \\
 \hline
\end{tabular}
\caption{Performance of different variations of $\mathcal{G}$ and of selection approaches on the 9 targets knowledge (performance metrics are the same used in Table~\ref{tab:cnn_results}).}
\label{tab:ae_results}
\end{table}
Table~\ref{tab:ae_results} shows that the Concat baseline is outperformed by all other approaches on almost all target knowledge. It is in line with what Zamir~\etal~\cite{zamir2018taskonomy} observed and it  might be explained by the large dimensionality of the input of the Concat method and the limited size of some of the target domains. 
Then, it is interesting to see that the simple PCA method is nevertheless several times on par with the BT results, while $\mathcal{G}$ performances are better than BT results and often better than the BCT.
Thus, it validates the choice of a reduction of dimensionality applied on all embedding $h_i$, moreover illustrating the intuition developed in introduction: \emph{when the embeddings from which we want to transfer knowledge are correlated, there is a benefit to exploit these redundancies instead of discarding some information by block}, as done by the selection methods. 
Note that in contrast to the direct Concat baseline approach, $\mathcal{G}$ is able to extract a meaningful $h_G$ from the $h_i$ because of the large number of observed samples of $\mathcal{D}_{unsup}$.

\subsection{Impact of $h_G$ dimension}
\begin{figure}[tb]
    \centering
    \includegraphics[width=0.8\linewidth]{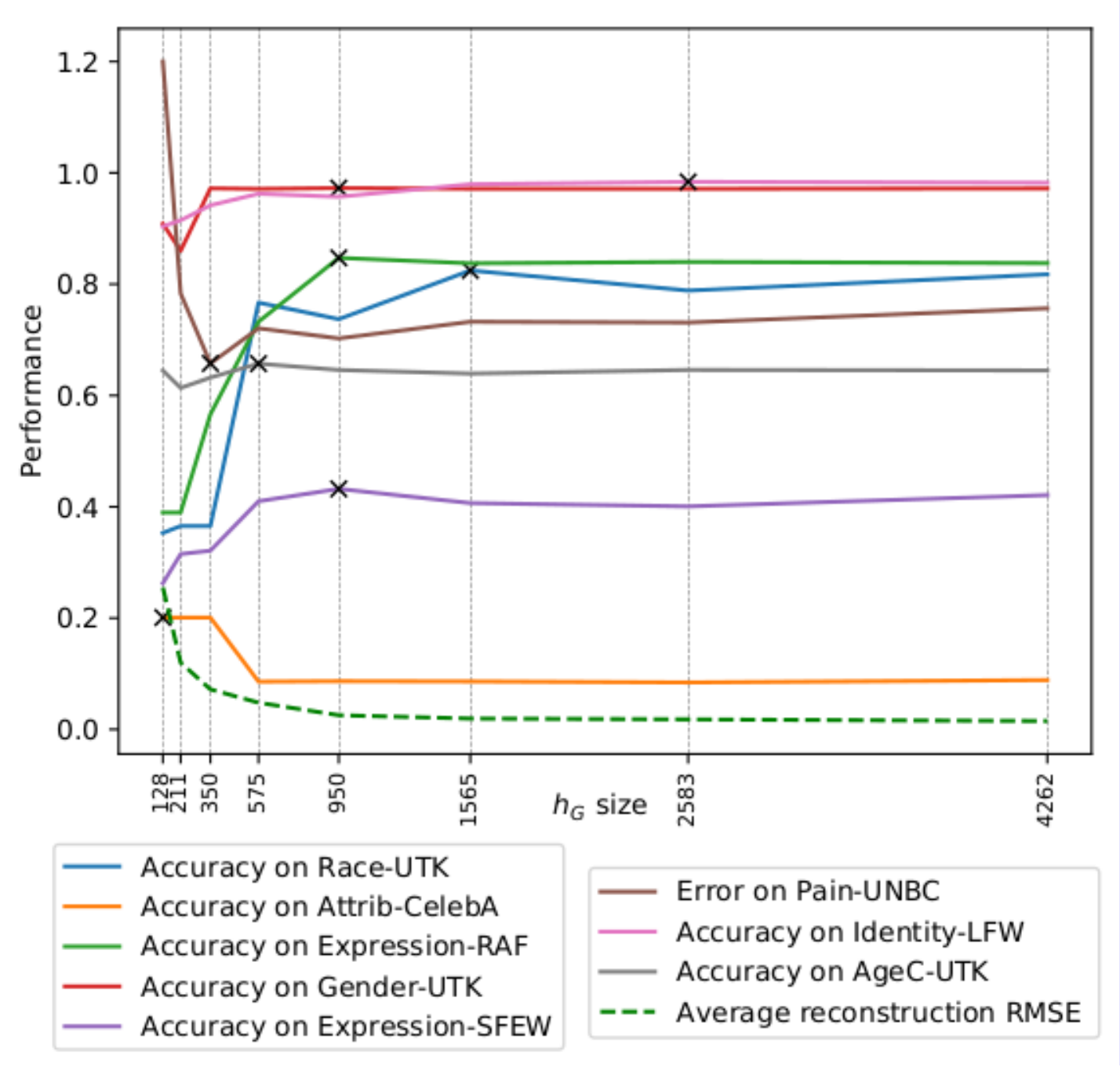}
    \caption{Impact of the size of $h_G$ on the reconstruction error of $\tilde{\mathcal{G}}$ and on the transfer for some of the target knowledge. Black cross is the optimal representation size for a given knowledge.}
    \label{fig:influence}
\end{figure}

During the presentation of the method, we propose to choose the size of $h_G$ as the average size of $h_i$. In Figure~\ref{fig:influence} we discuss this choice, by showing that there is no optimal representation size for all knowledge. Nevertheless, a too small representation conduct to low results and a large representation implies a high-computational cost and lower performance.
\label{rep_influence}

\subsection{Effect of Learning in 2 Stages}
During the construction of our model, we argue that there was a benefit to adopt a 2 stage approach, disentangling the compression and the distillation steps. We already have observed the low ability of $h_{MT}$ to reconstruct the source embedding $h_i$. Yet, what is the real impact in term of transfer learning of this insufficient convergence.

Let's study in details the columns dedicated to $\mathcal{E}^G_{MT}$ and $\mathcal{E}^G_{unique}$ in Table~\ref{tab:cnn_results}. We observe that if we are not fine-tuning the whole parameters of the model and only using the $h_{MT}$ and $\hat{h}_{G}$ as features, the gap is significant on most of the source and target knowledge. Moreover, these results are correlated task by task to the reconstruction error observed on each $h_i$.
Nevertheless $h_{MT}$ allows to achieve far better performance than the CNN baseline, still underlining the benefit of a pre-training, even when it is a noisy one.  

\subsection{Training Time}
\begin{figure}[tb]
    \centering
    \includegraphics[width=0.9\linewidth]{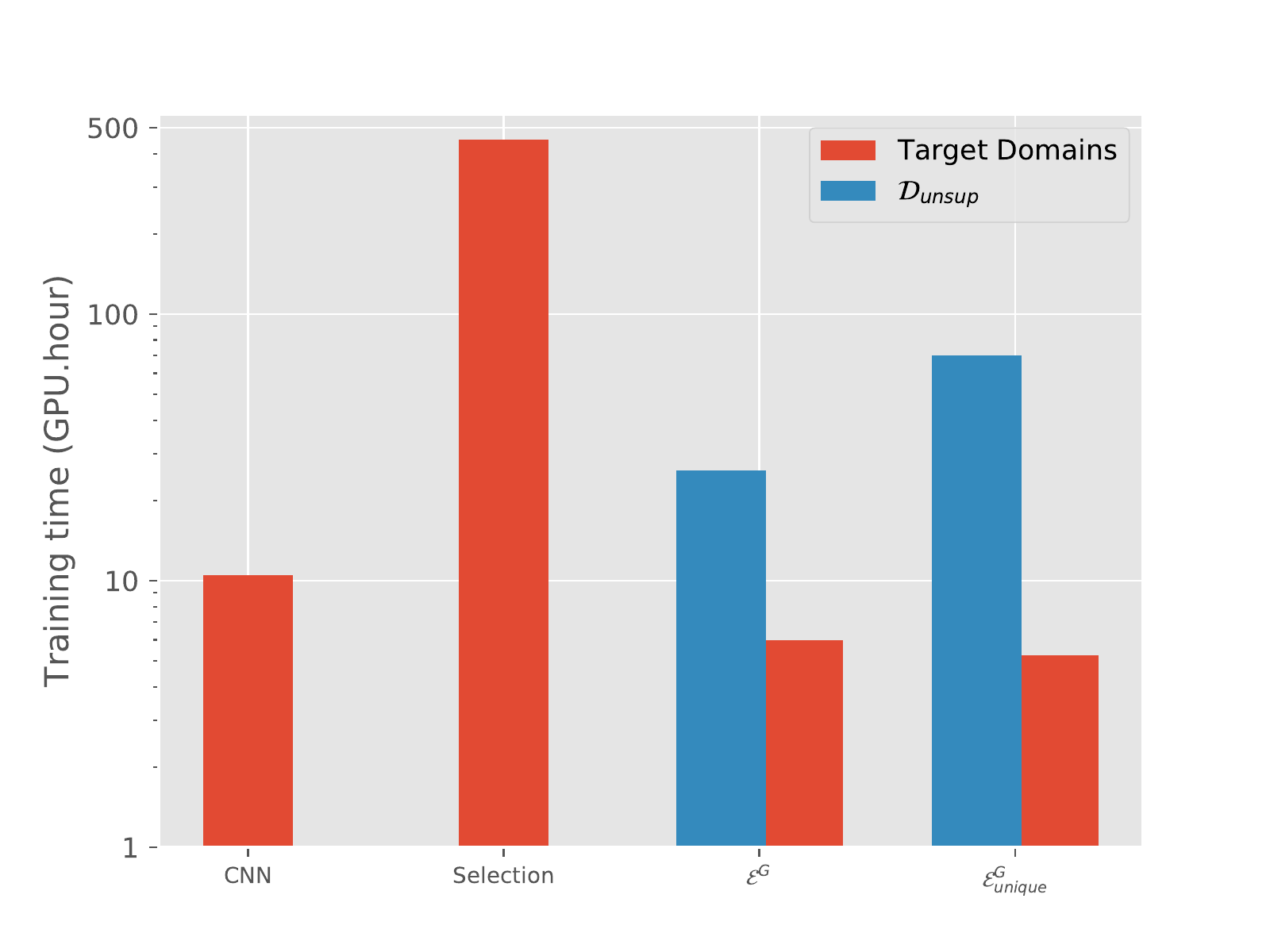}
    \caption{Total training time in hours (logscale) with one P100 GPU for different methods, on $\mathcal{D}_{unsup}$ and on all target domains.}
    \label{fig:times}
\end{figure}

Finally, we report from Figure~\ref{fig:times} the training time spent on $\mathcal{D}_{unsup}$ and on the tartget domains for different approaches. Even if the training on $\mathcal{D}_{unsup}$ is relatively long (almost 70 hours), the transfer learning step is then really faster than training a model from scratch. Moreover compared to the brute-force baseline of selecting all possible combinations (BCT), the total training time of the $\mathcal{E}^G_{unique}$ (taking the training time of $\mathcal{G}$ into account) is divided by 4. We can also project ourselves in the case where we had much more knowledge to master. For instance, multiplying the number of target knowledge by 2 will imply a factor of time of 8 between $\mathcal{E}^G_{unique}$ and BCT. 

\section{Conclusions}
This paper has introduced a novel approach to the problem of multi-source transfer learning, validated in the context of facial analysis. A unique and general model $\mathcal{E}^G_{unique}$, obtained by merging six different source knowledge, can be transferred on 9 different target knowledge.
Building this model is done through two successive training steps.
First, an autoencoder $\mathcal{G}$ if trained to combine the hidden representations of the existing models into one single unifying embedding $h_G$. 
Then, distilling this model to a light-weight student CNN allows to reduce the number of parameters and improve the adaptation ability of the model. The approach was experimentally validated by an exhaustive ablation study and performances on par with state-of-the-art methods on the 15 different knowledge, with a single simple model.  
~
On overall, the approach provides an efficient way to obtain universal models compressing the knowledge included in several existing models, without loss in performance, allowing an easy exploitation in real-world applications.


\bibliographystyle{ieee}
\bibliography{egbib}

\begin{thebibliography}{10}\itemsep=-1pt

\bibitem{acharya2018covariance}
D.~Acharya, Z.~Huang, D.~Pani~Paudel, and L.~Van~Gool.
\newblock Covariance pooling for facial expression recognition.
\newblock In {\em CVPR Workshop}, 2018.

\bibitem{achille2019task2vec}
A.~Achille, M.~Lam, R.~Tewari, A.~Ravichandran, S.~Maji, C.~Fowlkes, S.~Soatto,
  and P.~Perona.
\newblock Task2vec: Task embedding for meta-learning.
\newblock {\em arXiv preprint arXiv:1902.03545}, 2019.

\bibitem{andrew2013deep}
G.~Andrew, R.~Arora, J.~Bilmes, and K.~Livescu.
\newblock Deep canonical correlation analysis.
\newblock In {\em ICML}, 2013.

\bibitem{antipov2017effective}
G.~Antipov, M.~Baccouche, S.-A. Berrani, and J.-L. Dugelay.
\newblock Effective training of convolutional neural networks for face-based
  gender and age prediction.
\newblock {\em Pattern Recognition}, 2017.

\bibitem{atrey2010multimodal}
P.~K. Atrey, M.~A. Hossain, A.~El~Saddik, and M.~S. Kankanhalli.
\newblock Multimodal fusion for multimedia analysis: a survey.
\newblock {\em Multimedia systems}, 2010.

\bibitem{aytar2016soundnet}
Y.~Aytar, C.~Vondrick, and A.~Torralba.
\newblock Soundnet: Learning sound representations from unlabeled video.
\newblock In {\em NIPS}, 2016.

\bibitem{ballard1987modular}
D.~H. Ballard.
\newblock Modular learning in neural networks.
\newblock In {\em AAAI}, 1987.

\bibitem{baltruvsaitis2019multimodal}
T.~Baltru{\v{s}}aitis, C.~Ahuja, and L.-P. Morency.
\newblock Multimodal machine learning: A survey and taxonomy.
\newblock {\em T-PAMI}, 2019.

\bibitem{cao2018partially}
J.~Cao, Y.~Li, and Z.~Zhang.
\newblock Partially shared multi-task convolutional neural network with local
  constraint for face attribute learning.
\newblock In {\em CVPR}, 2018.

\bibitem{cao2018vggface2}
Q.~Cao, L.~Shen, W.~Xie, O.~M. Parkhi, and A.~Zisserman.
\newblock Vggface2: A dataset for recognising faces across pose and age.
\newblock In {\em FG}, 2018.

\bibitem{cao2019consistent}
W.~Cao, V.~Mirjalili, and S.~Raschka.
\newblock Consistent rank logits for ordinal regression with convolutional
  neural networks.
\newblock {\em arXiv preprint arXiv:1901.07884}, 2019.

\bibitem{cao2018partial}
Z.~Cao, M.~Long, J.~Wang, and M.~I. Jordan.
\newblock Partial transfer learning with selective adversarial networks.
\newblock In {\em CVPR}, 2018.

\bibitem{caron2018deep}
M.~Caron, P.~Bojanowski, A.~Joulin, and M.~Douze.
\newblock Deep clustering for unsupervised learning of visual features.
\newblock In {\em ECCV}, 2018.

\bibitem{chandar2016correlational}
S.~Chandar, M.~M. Khapra, H.~Larochelle, and B.~Ravindran.
\newblock Correlational neural networks.
\newblock {\em Neural computation}, 2016.

\bibitem{chang2019deep}
F.-J. Chang, A.~T. Tran, T.~Hassner, I.~Masi, R.~Nevatia, and G.~Medioni.
\newblock Deep, landmark-free fame: Face alignment, modeling, and expression
  estimation.
\newblock {\em IJCV}, 2019.

\bibitem{chebotar2016distilling}
Y.~Chebotar and A.~Waters.
\newblock Distilling knowledge from ensembles of neural networks for speech
  recognition.
\newblock In {\em Interspeech}, pages 3439--3443, 2016.

\bibitem{chen2018coupled}
S.~Chen, C.~Zhang, and M.~Dong.
\newblock Coupled end-to-end transfer learning with generalized fisher
  information.
\newblock In {\em CVPR}, 2018.

\bibitem{cheng2017survey}
Y.~Cheng, D.~Wang, P.~Zhou, and T.~Zhang.
\newblock A survey of model compression and acceleration for deep neural
  networks.
\newblock {\em arXiv preprint arXiv:1710.09282}, 2017.

\bibitem{chu2016best}
B.~Chu, V.~Madhavan, O.~Beijbom, J.~Hoffman, and T.~Darrell.
\newblock Best practices for fine-tuning visual classifiers to new domains.
\newblock In {\em ECCV Workshop}, 2016.

\bibitem{das2018mitigating}
A.~Das, A.~Dantcheva, and F.~Bremond.
\newblock Mitigating bias in gender, age and ethnicity classification: A
  multi-task convolution neural network approach.
\newblock In {\em ECCV Workshop}, 2018.

\bibitem{deng2009imagenet}
J.~Deng, W.~Dong, R.~Socher, L.-J. Li, K.~Li, and L.~Fei-Fei.
\newblock Imagenet: A large-scale hierarchical image database.
\newblock In {\em CVPR}, 2009.

\bibitem{sfew}
A.~Dhall, R.~Goecke, S.~Lucey, and T.~Gedeon.
\newblock Static facial expression analysis in tough conditions: Data,
  evaluation protocol and benchmark.
\newblock In {\em ICCV Workshop}, 2011.

\bibitem{doersch2017multi}
C.~Doersch and A.~Zisserman.
\newblock Multi-task self-supervised visual learning.
\newblock In {\em ICCV}, 2017.

\bibitem{fabian2016emotionet}
C.~Fabian Benitez-Quiroz, R.~Srinivasan, and A.~M. Martinez.
\newblock Emotionet: An accurate, real-time algorithm for the automatic
  annotation of a million facial expressions in the wild.
\newblock In {\em CVPR}, 2016.

\bibitem{geyer2018transfer}
R.~C. Geyer, V.~Wegmayr, and L.~Corinzia.
\newblock Transfer learning by adaptive merging of multiple models.
\newblock In {\em MIDL}, 2019.

\bibitem{ghifary2015domain}
M.~Ghifary, W.~Bastiaan~Kleijn, M.~Zhang, and D.~Balduzzi.
\newblock Domain generalization for object recognition with multi-task
  autoencoders.
\newblock In {\em ICCV}, 2015.

\bibitem{gidaris2018unsupervised}
S.~Gidaris, P.~Singh, and N.~Komodakis.
\newblock Unsupervised representation learning by predicting image rotations.
\newblock In {\em ICLR}, 2018.

\bibitem{guo2016ms}
Y.~Guo, L.~Zhang, Y.~Hu, X.~He, and J.~Gao.
\newblock Ms-celeb-1m: A dataset and benchmark for large-scale face
  recognition.
\newblock In {\em ECCV}, 2016.

\bibitem{han2013age}
H.~Han, C.~Otto, and A.~K. Jain.
\newblock Age estimation from face images: Human vs. machine performance.
\newblock In {\em ICB}, 2013.

\bibitem{hardoon2004canonical}
D.~R. Hardoon, S.~Szedmak, and J.~Shawe-Taylor.
\newblock Canonical correlation analysis: An overview with application to
  learning methods.
\newblock {\em Neural computation}, 16(12):2639--2664, 2004.

\bibitem{he2016deep}
K.~He, X.~Zhang, S.~Ren, and J.~Sun.
\newblock Deep residual learning for image recognition.
\newblock In {\em CVPR}, 2016.

\bibitem{HintonVD15}
G.~E. Hinton, O.~Vinyals, and J.~Dean.
\newblock Distilling the knowledge in a neural network.
\newblock {\em arXiv preprint arXiv:1503.02531}, 2015.

\bibitem{lfw}
G.~B. Huang, M.~Ramesh, T.~Berg, and E.~Learned-Miller.
\newblock Labeled faces in the wild: A database for studying face recognition
  in unconstrained environments.
\newblock Technical Report 07-49, University of Massachusetts, Amherst, October
  2007.

\bibitem{huo2007survey}
X.~Huo, X.~S. Ni, and A.~K. Smith.
\newblock A survey of manifold-based learning methods.
\newblock {\em Recent advances in data mining of enterprise data}, 2007.

\bibitem{joliffe1992principal}
I.~Joliffe and B.~Morgan.
\newblock Principal component analysis and exploratory factor analysis.
\newblock {\em Statistical methods in medical research}, 1(1):69--95, 1992.

\bibitem{karras2017progressive}
T.~Karras, T.~Aila, S.~Laine, and J.~Lehtinen.
\newblock Progressive growing of gans for improved quality, stability, and
  variation.
\newblock In {\em ICLR}, 2018.

\bibitem{kingma2013auto}
D.~P. Kingma and M.~Welling.
\newblock Auto-encoding variational bayes.
\newblock In {\em ICLR}, 2014.

\bibitem{fgnet}
A.~Lanitis and T.~Cootes.
\newblock Fg-net aging data base.
\newblock {\em Cyprus College}, 2(3):5, 2002.

\bibitem{lee2017overcoming}
S.-W. Lee, J.-H. Kim, J.~Jun, J.-W. Ha, and B.-T. Zhang.
\newblock Overcoming catastrophic forgetting by incremental moment matching.
\newblock In {\em NIPS}, 2017.

\bibitem{li2019task}
J.~Li, P.~Zhou, Y.~Chen, J.~Zhao, S.~Roy, Y.~Shuicheng, J.~Feng, and T.~Sim.
\newblock Task relation networks.
\newblock In {\em WACV}, 2019.

\bibitem{li2018measure}
S.~Li.
\newblock Measure, manifold, learning, and optimization: A theory of neural
  networks.
\newblock {\em arXiv preprint arXiv:1811.12783}, 2018.

\bibitem{raf}
S.~Li, W.~Deng, and J.~Du.
\newblock Reliable crowdsourcing and deep locality-preserving learning for
  expression recognition in the wild.
\newblock In {\em CVPR}, 2017.

\bibitem{li2019delta}
X.~Li, H.~Xiong, H.~Wang, Y.~Rao, L.~Liu, and J.~Huan.
\newblock Delta: Deep learning transfer using feature map with attention for
  convolutional networks.
\newblock In {\em ICLR}, 2019.

\bibitem{celebA}
Z.~Liu, P.~Luo, X.~Wang, and X.~Tang.
\newblock Deep learning face attributes in the wild.
\newblock In {\em ICCV}, 2015.

\bibitem{pain}
P.~Lucey, J.~F. Cohn, K.~M. Prkachin, P.~E. Solomon, and I.~Matthews.
\newblock Painful data: The unbc-mcmaster shoulder pain expression archive
  database.
\newblock In {\em FG}, 2011.

\bibitem{mancini2018best}
M.~Mancini, S.~R. Bul{\`o}, B.~Caputo, and E.~Ricci.
\newblock Best sources forward: domain generalization through source-specific
  nets.
\newblock In {\em ICIP}, 2018.

\bibitem{mollahosseini2017affectnet}
A.~Mollahosseini, B.~Hasani, and M.~H. Mahoor.
\newblock Affectnet: A database for facial expression, valence, and arousal
  computing in the wild.
\newblock {\em Transactions on Affective Computing}, 2017.

\bibitem{neverova2016moddrop}
N.~Neverova, C.~Wolf, G.~Taylor, and F.~Nebout.
\newblock Moddrop: adaptive multi-modal gesture recognition.
\newblock {\em T-PAMI}, 2016.

\bibitem{ngiam2011multimodal}
J.~Ngiam, A.~Khosla, M.~Kim, J.~Nam, H.~Lee, and A.~Y. Ng.
\newblock Multimodal deep learning.
\newblock In {\em ICML}, 2011.

\bibitem{noroozi2016unsupervised}
M.~Noroozi and P.~Favaro.
\newblock Unsupervised learning of visual representations by solving jigsaw
  puzzles.
\newblock In {\em ECCV}, 2016.

\bibitem{noroozi2017representation}
M.~Noroozi, H.~Pirsiavash, and P.~Favaro.
\newblock Representation learning by learning to count.
\newblock In {\em ICCV}, 2017.

\bibitem{papernot2016semi}
N.~Papernot, M.~Abadi, U.~Erlingsson, I.~Goodfellow, and K.~Talwar.
\newblock Semi-supervised knowledge transfer for deep learning from private
  training data.
\newblock {\em ICLR}, 2017.

\bibitem{perez2019mfas}
J.-M. P{\'e}rez-R{\'u}a, V.~Vielzeuf, S.~Pateux, M.~Baccouche, and F.~Jurie.
\newblock Mfas: Multimodal fusion architecture search.
\newblock In {\em CVPR}, 2019.

\bibitem{radenovic2018deep}
F.~Radenovic, G.~Tolias, and O.~Chum.
\newblock Deep shape matching.
\newblock In {\em ECCV}, 2018.

\bibitem{radosavovic2018data}
I.~Radosavovic, P.~Doll{\'a}r, R.~Girshick, G.~Gkioxari, and K.~He.
\newblock Data distillation: Towards omni-supervised learning.
\newblock In {\em CVPR}, 2018.

\bibitem{romero2014fitnets}
A.~Romero, N.~Ballas, S.~E. Kahou, A.~Chassang, C.~Gatta, and Y.~Bengio.
\newblock Fitnets: Hints for thin deep nets.
\newblock In {\em ICLR}, 2015.

\bibitem{imdbwiki}
R.~Rothe, R.~Timofte, and L.~Van~Gool.
\newblock Dex: Deep expectation of apparent age from a single image.
\newblock In {\em ICCV Workshop}, 2015.

\bibitem{rothe2018deep}
R.~Rothe, R.~Timofte, and L.~Van~Gool.
\newblock Deep expectation of real and apparent age from a single image without
  facial landmarks.
\newblock {\em IJCV}, 2018.

\bibitem{ruder122019latent}
S.~Ruder, J.~Bingel, I.~Augenstein, and A.~S{\o}gaard.
\newblock Latent multi-task architecture learning.
\newblock In {\em AAAI}, 2019.

\bibitem{schroff2015facenet}
F.~Schroff, D.~Kalenichenko, and J.~Philbin.
\newblock Facenet: A unified embedding for face recognition and clustering.
\newblock In {\em CVPR}, 2015.

\bibitem{shahroudy2018deep}
A.~Shahroudy, T.-T. Ng, Y.~Gong, and G.~Wang.
\newblock Deep multimodal feature analysis for action recognition in rgb+ d
  videos.
\newblock {\em T-PAMI}, 2018.

\bibitem{snoek2005early}
C.~G. Snoek, M.~Worring, and A.~W. Smeulders.
\newblock Early versus late fusion in semantic video analysis.
\newblock In {\em ACMMM}, 2005.

\bibitem{vielzeuf2018occam}
V.~Vielzeuf, C.~Kervadec, S.~Pateux, A.~Lechervy, and F.~Jurie.
\newblock An occam's razor view on learning audiovisual emotion recognition
  with small training sets.
\newblock In {\em ICMI}, 2018.

\bibitem{vincent2010stacked}
P.~Vincent, H.~Larochelle, I.~Lajoie, Y.~Bengio, and P.-A. Manzagol.
\newblock Stacked denoising autoencoders: Learning useful representations in a
  deep network with a local denoising criterion.
\newblock {\em JMLR}, 2010.

\bibitem{xu2018pad}
D.~Xu, W.~Ouyang, X.~Wang, and N.~Sebe.
\newblock Pad-net: multi-tasks guided prediction-and-distillation network for
  simultaneous depth estimation and scene parsing.
\newblock In {\em CVPR}, 2018.

\bibitem{yao2015tiny}
L.~Yao and J.~Miller.
\newblock Tiny imagenet classification with convolutional neural networks.
\newblock {\em CS 231N}, 2015.

\bibitem{zamir2018taskonomy}
A.~R. Zamir, A.~Sax, W.~Shen, L.~J. Guibas, J.~Malik, and S.~Savarese.
\newblock Taskonomy: Disentangling task transfer learning.
\newblock In {\em CVPR}, 2018.

\bibitem{zeng2018facial}
J.~Zeng, S.~Shan, and X.~Chen.
\newblock Facial expression recognition with inconsistently annotated datasets.
\newblock In {\em ECCV}, 2018.

\bibitem{zhang2016colorful}
R.~Zhang, P.~Isola, and A.~A. Efros.
\newblock Colorful image colorization.
\newblock In {\em ECCV}, 2016.

\bibitem{zhang2017split}
R.~Zhang, P.~Isola, and A.~A. Efros.
\newblock Split-brain autoencoders: Unsupervised learning by cross-channel
  prediction.
\newblock In {\em CVPR}, 2017.

\bibitem{utk}
S.-Y. Zhang, Zhifei and H.~Qi.
\newblock Age progression/regression by conditional adversarial autoencoder.
\newblock In {\em CVPR}, 2017.

\bibitem{zhang2018bilateral}
Y.~Zhang, R.~Zhao, W.~Dong, B.-G. Hu, and Q.~Ji.
\newblock Bilateral ordinal relevance multi-instance regression for facial
  action unit intensity estimation.
\newblock In {\em CVPR}, 2018.

\end{thebibliography}

\end{document}